\documentclass{article}
\pdfoutput=1

\PassOptionsToPackage{numbers, compress}{natbib}

\usepackage[final]{nips_2018}

\usepackage[utf8]{inputenc} 
\usepackage[T1]{fontenc}    
\usepackage{hyperref}       
\usepackage{url}            
\usepackage{booktabs}       
\usepackage{amsfonts}       
\usepackage{nicefrac}       
\usepackage{microtype}      
\usepackage{graphicx}       
\usepackage{siunitx}
\usepackage{booktabs}
\usepackage{etoolbox}
\usepackage{subcaption}  

\title{Improving Reconstruction Autoencoder Out-of-distribution Detection with Mahalanobis Distance}

\author{
    Taylor~Denouden\thanks{Department of Computer Science, University of Waterloo} \\
    \texttt{tadenoud@uwaterloo.ca} \\
    \And
    Rick~Salay\thanks{Department of Electrical and Computer Engineering, University of Waterloo} \\
    \texttt{rsalay@gsd.uwaterloo.ca} \\
    \And
    Krzysztof~Czarnecki\footnotemark[2] \\
    \texttt{kczarnec@gsd.uwaterloo.ca} \\
    \And
    Vahdat~Abdelzad\footnotemark[2] \\
    \texttt{vabdelza@uwaterloo.ca} \\
    \And
    Buu~Phan\footnotemark[2] \\
    \texttt{btphan@uwaterloo.ca} \\
    \And
    Sachin~Vernekar\footnotemark[1] \\
    \texttt{sverneka@uwaterloo.ca} \\
}

\begin{document}

\maketitle  

\begin{abstract}
    There is an increasingly apparent need for validating the classifications made by deep learning systems in safety-critical applications like autonomous vehicle systems. A number of recent papers have proposed methods for detecting anomalous image data that appear different from known inlier data samples, including reconstruction-based autoencoders. Autoencoders optimize the compression of input data to a latent space of a dimensionality smaller than the original input and attempt to accurately reconstruct the input using that compressed representation. Since the latent vector is optimized to capture the salient features from the inlier class only, it is commonly assumed that images of objects from outside of the training class cannot effectively be compressed and reconstructed. Some thus consider reconstruction error as a kind of novelty measure. Here we suggest that reconstruction-based approaches fail to capture particular anomalies that lie far from known inlier samples in latent space but near the latent dimension manifold defined by the parameters of the model. We propose incorporating the Mahalanobis distance in latent space to better capture these out-of-distribution samples and our results show that this method often improves performance over the baseline approach.
\end{abstract}

\section{Introduction}
    Many computer vision applications operate in safety-critical environments where mistakes made by an autonomous agent have serious consequences. An example is autonomous vehicles that use perception systems to model the surrounding environment and help the vehicle avoid collisions. Many existing deep neural network classification models used by such perception systems cannot detect when a provided input is from an out-of-distribution (OOD) class with respect to the training data \cite{Hendrycks2017ANetworks}. As a result, there is uncertainty about how perception systems trained to detect specific objects will react when provided these unusual inputs. OOD detection algorithms attempt to rectify this issue by modifying or augmenting existing classification models in order to detect samples from a distribution different than the training dataset~\cite{Pimentel2014ADetection}.

    There are a variety of techniques for accomplishing this goal (see~\cite{Pimentel2014ADetection} for a survey), but this work will focus on reconstruction autoencoder-based approaches. This approach has the desirable characteristics of being able to operate in an unsupervised mode while making few assumptions about the underlying data-generating distribution of the inlier dataset~\cite{Pimentel2014ADetection, Amarbayasgalan2018UnsupervisedClustering}. In this paper, we attempt to improve autoencoder based method performance and show that reconstruction error alone is insufficient as a novelty metric. We then show that using distance metrics in the latent space of the autoencoder can help to address this problem and improve the performance of autoencoder OOD detectors with minimal additional computational cost.
    
\section{Related work}
    Previous work by \citet{Guo2018AnNeighbor} explored using KNN distance in the autoencoder latent space to improve OOD detection performance over using only the reconstruction error. While effective, this technique requires iterating over the training set at inference and subsequently has high computational and memory cost that makes it impractical for many real-world applications. \citet{An2015VariationalProbability} address this issue using variational autoencoders (VAE) and incorporate the probability of the latent embedding of the model in relation to some specified prior distribution. The VAE method improves OOD detection performance over vanilla autoencoders but requires making assumptions about the generating data distribution of the inlier class.
    
    Recent work by \citet{Lee2018AAttacks} proposed using Mahalanobis distance on the features learned by a deep classification model to detect OOD samples. Their method produced state-of-the-art results for multiple OOD detection benchmark datasets. Our work differs from \cite{Lee2018AAttacks} in that ours operates in an unsupervised mode where data labels are not required during training. Our method also allows doing OOD detection when only a single class of objects are considered to be the inlier class, which is not possible with classification-based methods.

\section{Background}
\subsection{Using autoencoders for OOD detection}
    In deep learning, autoencoders are a kind of deep neural network that are optimized to reconstruct the provided input data. These networks are typically designed with a hidden layer that is smaller in the number of dimensions than the original input, often called the \emph{bottleneck layer}, and are therefore optimized to learn a compressed representation of the input data distribution. As such, these models should be ineffective at compressing OOD images and unable to reconstruct the input without significant error~\cite{Pimentel2014ADetection, Sabokrou2018AdversariallyDetection}. The reconstruction error is thus considered to be a measure of novelty with respect to the inlier training dataset. Applying a threshold on this novelty score produces a binary classifier for detecting when a test sample is OOD.

\subsection{Mahalanobis distance}
    Mahalanobis distance ($D_{M}$) is a distance measure based on how many standard deviations a test sample $x$ is from the mean vector of a multivariate Gaussian distribution. It is defined as:
    
    \[D_{M}(x) = \sqrt{(x - \hat{\mu})^{T}\ \hat{\Sigma}^{-1}\ (x - \hat{\mu})}\]
    
    where $\hat{\mu}$ and $\hat{\Sigma}$ are the mean vector and covariance matrix of the distribution, respectively~\cite{Mahalanobis1936OnStatistics}. $D_{M}$ is scale-invariant and accounts for correlations between dimensions of the data used to estimate the distribution. This offers an advantage over Euclidean distance, which assumes that data dimensions are measured in terms of a common scale.

\section{Problem and approach}
    \begin{figure}
      \centering
      \includegraphics[width=320px,trim={0 0.6cm 0 0},clip]{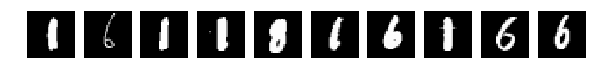}
      \vspace{1pt}
      \caption{MNIST OOD class digits with low reconstruction error. Model was trained with digit 0 images and has 16 bottleneck features. All digits shown have lower reconstruction error than 48\% or more of the test inlier data.}
      \label{fig:outlier-1-from-0}
    \end{figure}
    
    \begin{figure*}
        \centering
        \begin{subfigure}[t]{0.49\textwidth}
            \centering
            \includegraphics[height=1.4in,trim={5pt 0 7pt 0},clip]{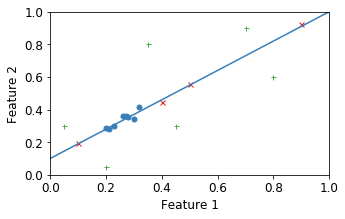}
            \caption{Linear manifold case}
            \label{fig:linear_problem_scatter}
        \end{subfigure}%
        ~ 
        \begin{subfigure}[t]{0.49\textwidth}
            \centering
            \includegraphics[height=1.4in,trim={5pt 0 5pt 0},clip]{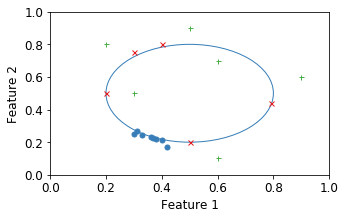}        
            \caption{Non-linear manifold case}
            \label{fig:non_linear_problem_scatter}
        \end{subfigure}
        \caption{Simplified reconstruction model OOD sample cases. The blue training data defines a latent subspace manifold (blue line in \emph{a} and circle in \emph{b}) that will minimize reconstruction error for those samples. OOD samples with high reconstruction error (shown as green +) will be detected but OOD samples with low reconstruction error (shown as red x) will not be.}
        \label{fig:problem_scatter}
    \end{figure*}
    
    Experimentally, we have found that autoencoders will sometimes reconstruct OOD samples with less error than many inlier samples. This behaviour is problematic because it results in a situation where it is impossible to set a novelty score threshold that correctly classifies these OOD samples while maintaining correct classification on all inlier samples with a comparatively higher reconstruction error. Figure~\ref{fig:outlier-1-from-0} shows various OOD examples from MNIST. We have observed autoencoders with varying architectures, such as dense and convolutional models with different depths and hidden layer sizes, exhibit this behavior, and we often encounter OOD samples with low reconstruction error that are visibly distinct from the inlier class to humans.
    
    We theorize that this issue occurs because some OOD samples lie near the latent dimension manifold defined by the bottleneck layer of the model but are mapped to a location on that manifold that is a significantly far from any known encoded training sample. Figure \ref{fig:problem_scatter} illustrates this scenario with examples of both linear and non-linear latent manifolds that exhibit this issue. A naive solution to this problem is to increase the number of latent dimensions to capture more of the variance of the original dataset. However, as the number of dimensions in the autoencoder increases so does the power of the model. An autoencoder with a bottleneck layer dimensionality equal to or greater than the original input can potentially learn the identity function and reconstruct any given input, diminishing its ability to distinguish between inlier and OOD data. Instead of increasing the size of the bottleneck layer, we propose using an additional measure of distance between the bottleneck layer latent embeddings of the training data and a given test sample. In doing this, we can account for data that is encoded to areas close to the latent space manifold yet far from the area where training samples are typically encoded to, as with the red x's Figure~\ref{fig:problem_scatter}.
    
\section{Experiments and results}
    \begin{figure*}
        \centering
        \begin{subfigure}[t]{0.49\textwidth}
            \centering
            \includegraphics[height=1.4in,trim={5pt 0 7pt 0},clip]{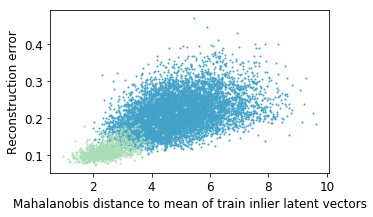}
            \caption{Light green inliers and blue OOD test samples plotted in terms of reconstruction error and latent space distance to the mean of the training inliers. Dataset is MNIST with digit 0 as the inliers. Model shown has 16 bottleneck features.}
            \label{fig:recon_dist_scatter}
        \end{subfigure}%
        ~ 
        \begin{subfigure}[t]{0.49\textwidth}
            \centering
            \includegraphics[height=1.4in,trim={5pt 0 5pt 0},clip]{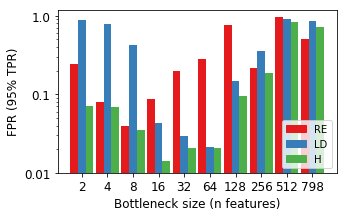}        
            \caption{OOD detection model performance on MNIST for inlier class 0 with varying bottleneck sizes. Lower FPR (95\% TPR) is better. Bars show results when using reconstruction error (RE), distance in latent space (LD), and a hybrid approach (H) as the novelty score.}
            \label{fig:mahalanobis_results}
        \end{subfigure}
        \caption{Visualization of data encoded by model and comparison of OOD detection results.}
    \end{figure*}
    
    We trained multiple reconstruction autoencoders with ten different bottleneck layer sizes varying between 2 and 798 (the original input size) to test our hypothesis. For each of the different architectures, a separate autoencoder was trained using only images from a single digit class from MNIST. The digit used for training was thereafter considered to be the inlier class for that particular model. More details regarding our training procedure are available in the appendix.
    
    To measure novelty on a test sample, we combined the Mahalanobis distance ($D_{M}$) between the encoded sample and the mean vector of the encoded training set with the reconstruction loss ($\ell$) of the sample as follows:
    
    \[novelty(x) = \alpha \cdot D_{M}\big{(} E(x)\big{)} + \beta \cdot \ell\big{(}x,\ D\big{(}E(x)\big{)}\big{)}\]
    
    where, $E$ and $D$ are the latent space encoding and decoding functions of the autoencoder, respectively, and $\alpha$ and $\beta$ are mixing parameters that were determined using a validation set of inliers samples. The $D_{M}$ was parameterized with $\hat{\mu}$ equal to the mean vector and $\hat{\Sigma}$ the covariance matrix of the encoded training data $E(X_{train})$.
    
    Mixing parameter $\alpha$ was set to the reciprocal of standard deviation of the Mahalanobis distance between the encoded validation data and the mean latent train vector and $\beta$ to the reciprocal of the standard deviation of the reconstruction error on the validation set. This normalization prevents either of these super-features from dominating the unified novelty score. Tuning this combination method may yield further performance improvements in future work.
    
    Our results show that incorporating the latent distance improved performance over using only the reconstruction error in most cases. Figure \ref{fig:recon_dist_scatter} shows a visualization of the test dataset plotted using these two super-features and illustrates how using the latent distance helps improve separation between the inlier and OOD data. Figure~\ref{fig:mahalanobis_results} shows the results for all autoencoder models trained with digit 0 as the inlier class in terms of the false positive rate when the true positive rate is set to 95\% (FPR (95\% TPR)), a standard OOD detection metric~\cite{Lee2018AAttacks, Liang2017EnhancingNetworks}. Table 1 compares the different methods for each of the inlier classes using established performance metrics from the OOD detection literature \citet{Hendrycks2017ANetworks, Liang2017EnhancingNetworks}. The bottleneck size that gives the best result using the hybrid novelty score for each digit varies between 8 and 64. Tables of all results are available in the appendix.

    \sisetup{detect-weight,mode=text}
    \renewrobustcmd{\bfseries}{\fontseries{b}\selectfont}
    \renewrobustcmd{\boldmath}{}
    \newrobustcmd{\B}{\bfseries}
    \addtolength{\tabcolsep}{-2pt}
    
    \begin{table}[!th]
      \caption{Results comparison between baseline reconstruction error OOD detection approach and our hybrid approach. Architecture chosen was for the best available result for the baseline method. The metrics shown are area under the receiver operating characteristic curve (AUROC), area under the precision recall curve when the in-distribution data is considered to be the positive class (AUPR (in)), area under the precision recall curve when the OOD data is considered to be the positive class (AUPR (out)), and false positive rate at 95\% true positive rate (FPR (95\% TPR)). Using a larger bottleneck and the hybrid score may also get better results than possible using reconstruction error alone. Best values are bolded.}
      \label{results-table}
      \centering
      \begin{tabular}{llllllllll}
        \toprule
        \multicolumn{2}{c}{ } & \multicolumn{2}{l}{FPR (95\% TPR)} & \multicolumn{2}{l}{AUROC} & \multicolumn{2}{l}{AUPR (in)} & \multicolumn{2}{l}{AUPR (out)} \\
        \cmidrule{3-4}
        \cmidrule{5-6}
        \cmidrule{7-8}
        \cmidrule{9-10}
        Inlier & Bottleneck & Recon. & Hybrid & Recon. & Hybrid & Recon. & Hybrid & Recon. & Hybrid \\
        class & size & err. & (ours) & err. & (ours) & err. & (ours) & err. & (ours) \\
        \midrule
        0 &	8 & 0.040 &	\B  0.035 &	0.989 &	\B	0.991 &	0.998 &	\B	0.999 &	0.051 &	\B	0.051 \\
        1 &	4 &	\B	0.004 &	0.007 &	\B	0.999 &	0.998 &	\B	1.000 &	1.000 &	\B	0.059 &	0.059 \\
        2 &	4 &	0.474 &	\B	0.335 &	0.900 &	\B	0.918 &	0.987 &	\B	0.988 &	\B	0.056 &	0.055 \\
        3 &	8 &	0.723 &	\B	0.413 &	0.861 &	\B	0.912 &	0.982 &	\B	0.988 &	\B	0.056 &	0.054 \\
        4 &	16 & 0.386 &	\B 0.258 &	0.932 &	\B	0.960 &	0.992 &	\B	0.995 &	0.052 &	\B	0.052 \\
        5 &	2 &	\B	0.272 &	0.656 &	\B	0.920 &	0.802 &	\B	0.990 &	0.975 &	0.047 &	\B	0.052 \\
        6 &	8 &	0.380 &	\B	0.280 &	0.945 &	\B	0.956 &	0.994 &	\B	0.995 &	\B	0.051 &	0.050 \\
        7 &	4 &	0.393 &	\B	0.209 &	0.932 &	\B	0.956 &	0.992 &	\B	0.994 &	\B	0.055 &	0.054 \\
        8 &	8 &	0.998 &	\B	0.625 &	0.778 &	\B	0.836 &	0.973 &	\B	0.978 &	\B	0.060 &	0.055 \\
        9 &	8 &	0.506 &	\B	0.263 &	0.916 &	\B	0.946 &	0.990 &	\B	0.993 &	\B	0.054 &	0.053 \\
        \bottomrule
      \end{tabular}
    \end{table}

\section{Conclusion and future work}
    In this work, we address the problem where autoencoders can accurately reconstruct image data from a class not present during training. We consider the case where OOD samples are embedded into the latent space far from the training data but in a location where they can be still be reconstructed with minimal error. To detect these cases we created autoencoders that used the Mahalanobis distance from the mean latent embedding of the training set in addition to the usual reconstruction error as a novelty score. Our results show that using this hybrid approach often improves performance over using reconstruction error alone.
    
    In the future, we would like to extend our work to more complex naturalistic image datasets, optimize the method in which latent distance and reconstruction error are combined, and generalize the method such that multiple image classes may be modeled as inliers by a single autoencoder. We would also like to investigate methods for optimizing the architecture of autoencoder models and experiment with applying our technique to denoising autoencoders for OOD detection.

\newpage
{\small
\bibliographystyle{IEEEtranN}
\bibliography{references}

\begin{thebibliography}{9}
\providecommand{\natexlab}[1]{#1}
\providecommand{\url}[1]{#1}
\csname url@samestyle\endcsname
\providecommand{\newblock}{\relax}
\providecommand{\bibinfo}[2]{#2}
\providecommand{\BIBentrySTDinterwordspacing}{\spaceskip=0pt\relax}
\providecommand{\BIBentryALTinterwordstretchfactor}{4}
\providecommand{\BIBentryALTinterwordspacing}{\spaceskip=\fontdimen2\font plus
\BIBentryALTinterwordstretchfactor\fontdimen3\font minus
  \fontdimen4\font\relax}
\providecommand{\BIBforeignlanguage}[2]{{%
\expandafter\ifx\csname l@#1\endcsname\relax
\typeout{** WARNING: IEEEtranN.bst: No hyphenation pattern has been}%
\typeout{** loaded for the language `#1'. Using the pattern for}%
\typeout{** the default language instead.}%
\else
\language=\csname l@#1\endcsname
\fi
#2}}
\providecommand{\BIBdecl}{\relax}
\BIBdecl

\bibitem[Hendrycks and Gimpel(2017)]{Hendrycks2017ANetworks}
\BIBentryALTinterwordspacing
D.~Hendrycks and K.~Gimpel, ``{A Baseline for Detecting Misclassified and
  Out-of-Distribution Examples in Neural Networks},'' in \emph{Proceedings of
  International Conference on Learning Representations}, 10 2017. [Online].
  Available: \url{http://arxiv.org/abs/1610.02136}
\BIBentrySTDinterwordspacing

\bibitem[Pimentel et~al.(2014)Pimentel, Clifton, Clifton, and
  Tarassenko]{Pimentel2014ADetection}
\BIBentryALTinterwordspacing
M.~A. Pimentel, D.~A. Clifton, L.~Clifton, and L.~Tarassenko, ``{A review of
  novelty detection},'' \emph{Signal Processing}, vol.~99, pp. 215--249, 2014.
  [Online]. Available: \url{http://dx.doi.org/10.1016/j.sigpro.2013.12.026}
\BIBentrySTDinterwordspacing

\bibitem[Amarbayasgalan et~al.(2018)Amarbayasgalan, Jargalsaikhan, Ryu,
  Amarbayasgalan, Jargalsaikhan, and
  Ryu]{Amarbayasgalan2018UnsupervisedClustering}
\BIBentryALTinterwordspacing
T.~Amarbayasgalan, B.~Jargalsaikhan, K.~Ryu, T.~Amarbayasgalan,
  B.~Jargalsaikhan, and K.~H. Ryu, ``{Unsupervised Novelty Detection Using Deep
  Autoencoders with Density Based Clustering},'' \emph{Applied Sciences},
  vol.~8, no.~9, p. 1468, 8 2018. [Online]. Available:
  \url{http://www.mdpi.com/2076-3417/8/9/1468}
\BIBentrySTDinterwordspacing

\bibitem[Guo et~al.(2018)Guo, Liu, Zuo, and Wu]{Guo2018AnNeighbor}
\BIBentryALTinterwordspacing
J.~Guo, G.~Liu, Y.~Zuo, and J.~Wu, ``{An Anomaly Detection Framework Based on
  Autoencoder and Nearest Neighbor},'' in \emph{2018 15th International
  Conference on Service Systems and Service Management (ICSSSM)}.\hskip 1em
  plus 0.5em minus 0.4em\relax IEEE, 7 2018, pp. 1--6. [Online]. Available:
  \url{https://ieeexplore.ieee.org/document/8464983/}
\BIBentrySTDinterwordspacing

\bibitem[An and Cho(2015)]{An2015VariationalProbability}
\BIBentryALTinterwordspacing
J.~An and S.~Cho, ``{Variational Autoencoder based Anomaly Detection using
  Reconstruction Probability},'' \emph{SNU Data Mining Center, Tech. Rep.}, pp.
  1--18, 2015. [Online]. Available:
  \url{https://www.semanticscholar.org/paper/Variational-Autoencoder-based-Anomaly-Detection-An-Cho/061146b1d7938d7a8dae70e3531a00fceb3c78e8}
\BIBentrySTDinterwordspacing

\bibitem[Lee et~al.(2018)Lee, Lee, Lee, and Shin]{Lee2018AAttacks}
\BIBentryALTinterwordspacing
K.~Lee, K.~Lee, H.~Lee, and J.~Shin, ``{A Simple Unified Framework for
  Detecting Out-of-Distribution Samples and Adversarial Attacks},''
  \emph{CoRR}, vol. abs/1807.0, 7 2018. [Online]. Available:
  \url{http://arxiv.org/abs/1807.03888}
\BIBentrySTDinterwordspacing

\bibitem[Sabokrou et~al.(2018)Sabokrou, Khalooei, Fathy, and
  Adeli]{Sabokrou2018AdversariallyDetection}
\BIBentryALTinterwordspacing
M.~Sabokrou, M.~Khalooei, M.~Fathy, and E.~Adeli, ``{Adversarially Learned
  One-Class Classifier for Novelty Detection},'' in \emph{IEEE Computer Society
  Conference on Computer Vision and Pattern Recognition}, 2018. [Online].
  Available:
  \url{http://openaccess.thecvf.com/content_cvpr_2018/papers/Sabokrou_Adversarially_Learned_One-Class_CVPR_2018_paper.pdf
  http://arxiv.org/abs/1802.09088}
\BIBentrySTDinterwordspacing

\bibitem[Mahalanobis(1936)]{Mahalanobis1936OnStatistics}
\BIBentryALTinterwordspacing
P.~Mahalanobis, ``{On the Generalized Distance in Statistics},'' pp. 49--55,
  1936. [Online]. Available:
  \url{http://library.isical.ac.in:8080/xmlui/handle/123456789/6765}
\BIBentrySTDinterwordspacing

\bibitem[Liang et~al.(2017)Liang, Li, and Srikant]{Liang2017EnhancingNetworks}
\BIBentryALTinterwordspacing
S.~Liang, Y.~Li, and R.~Srikant, ``{Enhancing The Reliability of
  Out-of-distribution Image Detection in Neural Networks},'' in
  \emph{Proceedings of International Conference on Learning Representations}, 6
  2017. [Online]. Available: \url{http://arxiv.org/abs/1706.02690}
\BIBentrySTDinterwordspacing

\end{thebibliography}
}

\newpage

\section*{Appendix}
\subsection*{Model architecture}

    The autoencoder architecture consists of two 3x3 convolutional layers with ReLU activation function, each followed by a max pooling operation. The first convolution had 32 filters, and the second had 2 filters. Following these convolutional layers, the output was flattened and passed through a dense layer with L1 regularization multiplied by a constant of 0.00001. The size of this dense layer was varied to investigate the effect of differently sized bottlenecks in the model. Bottleneck sizes 2, 4, 8, 16, 32, 64, 128, 256, 512, and 798 were tested. The decoder consisted again of 2 convolutional layers with ReLU activation functions, each followed by a 2x2 nearest-neighbor interpolation upsampling layer. The number of filters for these convolutional layers were 2 and 32 respectively. The final layer of the decoder was a single 3x3 convolutional layer with sigmoid activation. All convolutional layers used padding that was determined such that the input shape was preserved after the convolutional operation.

\subsection*{Training procedure}
    Each model was trained using an adadelta optimizer with binary cross-entropy loss. Binary cross-entropy loss was used because the image data was normalized to take values in the range [0, 1] and it was experimentally found to converge more quickly to a better solution than mean squared error. A random split was done on the MNIST training set to produce a validation dataset of size 10000 and a training set to 50000 samples. Each model was trained to a maximum of 500 epochs, with early stopping used if there was no improvement in the validation data loss for 20 epochs.

\subsection*{Complete results}
    \begin{table}[ht]
      \caption{Results for class 0 inlier between baseline reconstruction error OOD detection approach and our hybrid approach. Best values are bolded.}
      \centering
      \begin{tabular}{llllllllll}
        \toprule
        \multicolumn{2}{c}{ } & \multicolumn{2}{l}{FPR (95\% TPR)} & \multicolumn{2}{l}{AUROC} & \multicolumn{2}{l}{AUPR (in)} & \multicolumn{2}{l}{AUPR (out)} \\
        \cmidrule{3-4}
        \cmidrule{5-6}
        \cmidrule{7-8}
        \cmidrule{9-10}
        Inlier & Bottleneck & Recon. & Hybrid & Recon. & Hybrid & Recon. & Hybrid & Recon. & Hybrid \\
        class & size & err. & (ours) & err. & (ours) & err. & (ours) & err. & (ours) \\
        \midrule
        0 & 2   & 0.240 & \B 0.070 & 0.948 & \B 0.983 & 0.993 & \B 0.998 & 0.051 & 0.051 \\
        0 & 4   & 0.081 & \B 0.068 & 0.983 & \B 0.984 & 0.997 & \B 0.998 & 0.051 & 0.051 \\
        0 & 8   & 0.040 & \B 0.035 & 0.989 & \B 0.991 & 0.998 & \B 0.999 & 0.051 & 0.051 \\
        0 & 16  & 0.088 & \B 0.014 & 0.979 & \B 0.995 & 0.997 & \B 0.999 & 0.051 & 0.051 \\
        0 & 32  & 0.196 & \B 0.020 & 0.957 & \B 0.994 & 0.995 & \B 0.999 & 0.051 & 0.051 \\
        0 & 64  & 0.281 & \B 0.020 & 0.944 & \B 0.994 & 0.993 & \B 0.999 & \B 0.052 & 0.051 \\
        0 & 128 & 0.764 & \B 0.095 & 0.854 & \B 0.972 & 0.981 & \B 0.995 & \B 0.055 & 0.051 \\
        0 & 256 & 0.217 & \B 0.186 & 0.954 & \B 0.962 & 0.994 & \B 0.995 & 0.051 & 0.051 \\
        0 & 512 & 0.967 & \B 0.842 & \B 0.848 & 0.773 & \B 0.982 & 0.968 & 0.057 & \B 0.059 \\
        0 & 798 & \B 0.500 & 0.726 & \B 0.905 & 0.802 & \B 0.987 & 0.972 & 0.053 & \B 0.057 \\
        \bottomrule
      \end{tabular}
    \end{table}
    
    \begin{table}[ht]
      \caption{Results for class 1 inlier between baseline reconstruction error OOD detection approach and our hybrid approach. Best values are bolded.}
      \centering
      \begin{tabular}{llllllllll}
        \toprule
        \multicolumn{2}{c}{ } & \multicolumn{2}{l}{FPR (95\% TPR)} & \multicolumn{2}{l}{AUROC} & \multicolumn{2}{l}{AUPR (in)} & \multicolumn{2}{l}{AUPR (out)} \\
        \cmidrule{3-4}
        \cmidrule{5-6}
        \cmidrule{7-8}
        \cmidrule{9-10}
        Inlier & Bottleneck & Recon. & Hybrid & Recon. & Hybrid & Recon. & Hybrid & Recon. & Hybrid \\
        class & size & err. & (ours) & err. & (ours) & err. & (ours) & err. & (ours) \\
        1 & 2   & \B 0.006 & 0.010 & \B 0.998 & 0.997 & 1.000 & 1.000 & 0.059 & 0.059 \\
        1 & 4   & \B 0.004 & 0.007 & \B 0.999 & 0.998 & 1.000 & 1.000 & 0.059 & 0.059 \\
        1 & 8   & \B 0.005 & 0.006 & 0.999 & 0.999 & 1.000 & 1.000 & 0.059 & 0.059 \\
        1 & 16  & \B 0.006 & 0.008 & 0.998 & 0.998 & 1.000 & 1.000 & 0.059 & 0.059 \\
        1 & 32  & \B 0.004 & 0.012 & \B 0.998 & 0.997 & \B 1.000 & 0.999 & 0.059 & 0.059 \\
        1 & 64  & \B 0.007 & 0.010 & \B 0.998 & 0.997 & \B 1.000 & 0.999 & 0.059 & 0.059 \\
        1 & 128 & \B 0.006 & 0.020 & \B 0.999 & 0.992 & \B 1.000 & 0.997 & 0.059 & 0.059 \\
        1 & 256 & \B 0.004 & 0.012 & \B 0.999 & 0.995 & \B 1.000 & 0.999 & 0.059 & 0.059 \\
        1 & 512 & \B 0.007 & 0.023 & \B 0.998 & 0.993 & \B 1.000 & 0.999 & 0.059 & 0.059 \\
        1 & 798 & \B 0.004 & 0.023 & \B 0.998 & 0.994 & \B 1.000 & 0.999 & 0.059 & 0.059 \\
        \bottomrule
      \end{tabular}
    \end{table}
    
    \begin{table}[ht]
      \caption{Results for class 2 inlier between baseline reconstruction error OOD detection approach and our hybrid approach. Best values are bolded.}
      \centering
      \begin{tabular}{llllllllll}
        \toprule
        \multicolumn{2}{c}{ } & \multicolumn{2}{l}{FPR (95\% TPR)} & \multicolumn{2}{l}{AUROC} & \multicolumn{2}{l}{AUPR (in)} & \multicolumn{2}{l}{AUPR (out)} \\
        \cmidrule{3-4}
        \cmidrule{5-6}
        \cmidrule{7-8}
        \cmidrule{9-10}
        Inlier & Bottleneck & Recon. & Hybrid & Recon. & Hybrid & Recon. & Hybrid & Recon. & Hybrid \\
        class & size & err. & (ours) & err. & (ours) & err. & (ours) & err. & (ours) \\
        2 & 2   & 0.751 & \B 0.658 & \B 0.872 & 0.820 & \B 0.982 & 0.973 & 0.057 & \B 0.059 \\
        2 & 4   & 0.474 & \B 0.335 & 0.900 & \B 0.918 & 0.987 & \B 0.988 & \B 0.056 & 0.055 \\
        2 & 8   & 0.959 & \B 0.343 & 0.810 & \B 0.929 & 0.975 & \B 0.990 & \B 0.061 & 0.055 \\
        2 & 16  & 1.000 & \B 0.263 & 0.827 & \B 0.938 & 0.978 & \B 0.991 & \B 0.061 & 0.054 \\
        2 & 32  & 1.000 & \B 0.343 & 0.738 & \B 0.932 & 0.964 & \B 0.991 & \B 0.066 & 0.055 \\
        2 & 64  & 1.000 & \B 0.277 & 0.696 & \B 0.937 & 0.954 & \B 0.991 & \B 0.068 & 0.054 \\
        2 & 128 & 1.000 & \B 0.672 & 0.715 & \B 0.871 & 0.960 & \B 0.980 & \B 0.067 & 0.060 \\
        2 & 256 & 1.000 & \B 0.948 & 0.699 & \B 0.787 & 0.955 & \B 0.971 & \B 0.068 & 0.062 \\
        2 & 512 & 1.000 & \B 0.977 & \B 0.691 & 0.662 & \B 0.954 & 0.945 & 0.069 & \B 0.073 \\
        2 & 798 & 1.000 & \B 0.989 & \B 0.652 & 0.599 & \B 0.948 & 0.933 & 0.072 & \B 0.079 \\
        \bottomrule
      \end{tabular}
    \end{table}
    
    \begin{table}[ht]
      \caption{Results for class 3 inlier between baseline reconstruction error OOD detection approach and our hybrid approach. Best values are bolded.}
      \centering
      \begin{tabular}{llllllllll}
        \toprule
        \multicolumn{2}{c}{ } & \multicolumn{2}{l}{FPR (95\% TPR)} & \multicolumn{2}{l}{AUROC} & \multicolumn{2}{l}{AUPR (in)} & \multicolumn{2}{l}{AUPR (out)} \\
        \cmidrule{3-4}
        \cmidrule{5-6}
        \cmidrule{7-8}
        \cmidrule{9-10}
        Inlier & Bottleneck & Recon. & Hybrid & Recon. & Hybrid & Recon. & Hybrid & Recon. & Hybrid \\
        class & size & err. & (ours) & err. & (ours) & err. & (ours) & err. & (ours) \\
        3 & 2   & 0.968 & \B 0.382 & 0.798 & \B 0.841 & 0.974 & 0.974 & 0.060 & 0.056 \\
        3 & 4   & 0.956 & \B 0.369 & 0.804 & \B 0.896 & 0.975 & \B 0.985 & \B 0.060 & 0.054 \\
        3 & 8   & 0.723 & \B 0.413 & 0.861 & \B 0.912 & 0.982 & \B 0.988 & \B 0.056 & 0.054 \\
        3 & 16  & 0.838 & \B 0.426 & 0.833 & \B 0.920 & 0.978 & \B 0.990 & \B 0.058 & 0.054 \\
        3 & 32  & 0.990 & \B 0.438 & 0.739 & \B 0.908 & 0.964 & \B 0.988 & \B 0.064 & 0.054 \\
        3 & 64  & 0.972 & \B 0.260 & 0.775 & \B 0.932 & 0.971 & \B 0.990 & \B 0.062 & 0.053 \\
        3 & 128 & 0.999 & \B 0.540 & 0.765 & \B 0.875 & 0.969 & \B 0.982 & \B 0.063 & 0.055 \\
        3 & 256 & 0.987 & \B 0.891 & \B 0.769 & 0.734 & \B 0.969 & 0.962 & 0.062 & \B 0.064 \\
        3 & 512 & 0.997 & \B 0.883 & \B 0.760 & 0.741 & \B 0.968 & 0.962 & 0.063 & 0.063 \\
        3 & 798 & 0.993 & \B 0.957 & \B 0.727 & 0.672 & \B 0.961 & 0.952 & \B 0.065 & 0.069 \\
      \end{tabular}
    \end{table}
    
    \begin{table}[ht]
      \caption{Results for class 4 inlier between baseline reconstruction error OOD detection approach and our hybrid approach. Best values are bolded.}
      \centering
      \begin{tabular}{llllllllll}
        \toprule
        \multicolumn{2}{c}{ } & \multicolumn{2}{l}{FPR (95\% TPR)} & \multicolumn{2}{l}{AUROC} & \multicolumn{2}{l}{AUPR (in)} & \multicolumn{2}{l}{AUPR (out)} \\
        \cmidrule{3-4}
        \cmidrule{5-6}
        \cmidrule{7-8}
        \cmidrule{9-10}
        Inlier & Bottleneck & Recon. & Hybrid & Recon. & Hybrid & Recon. & Hybrid & Recon. & Hybrid \\
        class & size & err. & (ours) & err. & (ours) & err. & (ours) & err. & (ours) \\
        4 & 2   & 0.920 & \B 0.568 & 0.855 & \B 0.876 & 0.983 & \B 0.984 & \B 0.056 & 0.054 \\
        4 & 4   & 0.618 & \B 0.435 & 0.905 & \B 0.915 & 0.989 & \B 0.990 & \B 0.054 & 0.053 \\
        4 & 8   & 0.673 & \B 0.503 & 0.895 & \B 0.906 & 0.988 & \B 0.989 & \B 0.054 & 0.053 \\
        4 & 16  & 0.386 & \B 0.258 & 0.932 & \B 0.960 & 0.992 & \B 0.995 & 0.052 & 0.052 \\
        4 & 32  & 0.853 & \B 0.264 & 0.868 & \B 0.959 & 0.985 & \B 0.995 & \B 0.056 & 0.052 \\
        4 & 64  & 0.930 & \B 0.247 & 0.863 & \B 0.958 & 0.985 & \B 0.995 & \B 0.056 & 0.051 \\
        4 & 128 & 0.995 & \B 0.329 & 0.823 & \B 0.932 & 0.980 & \B 0.991 & \B 0.058 & 0.052 \\
        4 & 256 & 0.998 & \B 0.576 & 0.807 & \B 0.877 & 0.978 & \B 0.983 & \B 0.059 & 0.054 \\
        4 & 512 & 0.989 & \B 0.767 & \B 0.836 & 0.833 & \B 0.981 & 0.978 & \B 0.058 & 0.056 \\
        4 & 798 & 0.988 & \B 0.815 & \B 0.823 & 0.810 & \B 0.980 & 0.975 & \B 0.058 & 0.057 \\
        \bottomrule
      \end{tabular}
    \end{table}
    
    \begin{table}[ht]
      \caption{Results for class 5 inlier between baseline reconstruction error OOD detection approach and our hybrid approach. Best values are bolded.}
      \centering
      \begin{tabular}{llllllllll}
        \toprule
        \multicolumn{2}{c}{ } & \multicolumn{2}{l}{FPR (95\% TPR)} & \multicolumn{2}{l}{AUROC} & \multicolumn{2}{l}{AUPR (in)} & \multicolumn{2}{l}{AUPR (out)} \\
        \cmidrule{3-4}
        \cmidrule{5-6}
        \cmidrule{7-8}
        \cmidrule{9-10}
        Inlier & Bottleneck & Recon. & Hybrid & Recon. & Hybrid & Recon. & Hybrid & Recon. & Hybrid \\
        class & size & err. & (ours) & err. & (ours) & err. & (ours) & err. & (ours) \\
        5 & 2   & \B 0.272 & 0.656 & \B 0.920 & 0.802 & \B 0.990 & 0.975 & 0.047 & \B 0.052 \\
        5 & 4   & 0.537 & \B 0.408 & \B 0.904 & 0.875 & \B 0.989 & 0.984 & 0.048 & 0.048 \\
        5 & 8   & 0.982 & \B 0.383 & 0.855 & \B 0.911 & 0.985 & \B 0.989 & \B 0.051 & 0.047 \\
        5 & 16  & 0.941 & \B 0.337 & 0.846 & \B 0.922 & 0.983 & \B 0.991 & \B 0.051 & 0.047 \\
        5 & 32  & 0.998 & \B 0.202 & 0.807 & \B 0.944 & 0.979 & \B 0.993 & \B 0.053 & 0.047 \\
        5 & 64  & 0.989 & \B 0.221 & 0.790 & \B 0.942 & 0.976 & \B 0.992 & \B 0.054 & 0.047 \\
        5 & 128 & 1.000 & \B 0.457 & 0.732 & \B 0.885 & 0.969 & \B 0.984 & \B 0.057 & 0.048 \\
        5 & 256 & 1.000 & \B 0.873 & 0.777 & \B 0.778 & \B 0.975 & 0.972 & \B 0.055 & 0.054 \\
        5 & 512 & 1.000 & \B 0.861 & 0.751 & \B 0.755 & \B 0.971 & 0.969 & \B 0.056 & 0.055 \\
        5 & 798 & 0.997 & \B 0.933 & \B 0.731 & 0.635 & \B 0.968 & 0.949 & 0.057 & \B 0.064 \\
        \bottomrule
      \end{tabular}
    \end{table}
    
    \begin{table}[ht]
      \caption{Results for class 6 inlier between baseline reconstruction error OOD detection approach and our hybrid approach. Best values are bolded.}
      \centering
      \begin{tabular}{llllllllll}
        \toprule
        \multicolumn{2}{c}{ } & \multicolumn{2}{l}{FPR (95\% TPR)} & \multicolumn{2}{l}{AUROC} & \multicolumn{2}{l}{AUPR (in)} & \multicolumn{2}{l}{AUPR (out)} \\
        \cmidrule{3-4}
        \cmidrule{5-6}
        \cmidrule{7-8}
        \cmidrule{9-10}
        Inlier & Bottleneck & Recon. & Hybrid & Recon. & Hybrid & Recon. & Hybrid & Recon. & Hybrid \\
        class & size & err. & (ours) & err. & (ours) & err. & (ours) & err. & (ours) \\
        6 & 2   & 0.524 & \B 0.288 & \B 0.938 & 0.928 & \B 0.993 & 0.991 & \B 0.051 & 0.050 \\
        6 & 4   & 0.570 & \B 0.328 & 0.927 & \B 0.932 & \B 0.992 & 0.991 & \B 0.052 & 0.050 \\
        6 & 8   & 0.380 & \B 0.280 & 0.945 & \B 0.956 & 0.994 & \B 0.995 & \B 0.051 & 0.050 \\
        6 & 16  & 0.703 & \B 0.148 & 0.904 & \B 0.977 & 0.989 & \B 0.997 & \B 0.052 & 0.050 \\
        6 & 32  & 0.836 & \B 0.058 & 0.900 & \B 0.989 & 0.989 & \B 0.999 & \B 0.053 & 0.050 \\
        6 & 64  & 0.934 & \B 0.053 & 0.882 & \B 0.990 & 0.987 & \B 0.999 & \B 0.054 & 0.050 \\
        6 & 128 & 0.916 & \B 0.115 & 0.867 & \B 0.978 & 0.985 & \B 0.997 & \B 0.054 & 0.050 \\
        6 & 256 & 0.878 & \B 0.418 & 0.888 & \B 0.926 & 0.987 & \B 0.991 & \B 0.053 & 0.051 \\
        6 & 512 & 0.970 & \B 0.747 & 0.856 & \B 0.873 & 0.984 & \B 0.985 & \B 0.055 & 0.053 \\
        6 & 798 & 0.903 & \B 0.674 & 0.858 & \B 0.876 & 0.983 & \B 0.985 & \B 0.054 & 0.053 \\
        \bottomrule
      \end{tabular}
    \end{table}
    
    \begin{table}[ht]
      \caption{Results for class 7 inlier between baseline reconstruction error OOD detection approach and our hybrid approach. Best values are bolded.}
      \centering
      \begin{tabular}{llllllllll}
        \toprule
        \multicolumn{2}{c}{ } & \multicolumn{2}{l}{FPR (95\% TPR)} & \multicolumn{2}{l}{AUROC} & \multicolumn{2}{l}{AUPR (in)} & \multicolumn{2}{l}{AUPR (out)} \\
        \cmidrule{3-4}
        \cmidrule{5-6}
        \cmidrule{7-8}
        \cmidrule{9-10}
        Inlier & Bottleneck & Recon. & Hybrid & Recon. & Hybrid & Recon. & Hybrid & Recon. & Hybrid \\
        class & size & err. & (ours) & err. & (ours) & err. & (ours) & err. & (ours) \\
        7 & 2   & 0.688 & \B 0.294 & 0.906 & \B 0.936 & 0.989 & \B 0.992 & \B 0.056 & 0.054 \\
        7 & 4   & 0.393 & \B 0.209 & 0.932 & \B 0.956 & 0.992 & \B 0.994 & \B 0.055 & 0.054 \\
        7 & 8   & 0.498 & \B 0.233 & 0.926 & \B 0.954 & 0.991 & \B 0.994 & \B 0.055 & 0.054 \\
        7 & 16  & 0.665 & \B 0.202 & 0.910 & \B 0.960 & 0.989 & \B 0.995 & \B 0.056 & 0.054 \\
        7 & 32  & 0.767 & \B 0.198 & 0.901 & \B 0.964 & 0.988 & \B 0.995 & \B 0.057 & 0.054 \\
        7 & 64  & 0.954 & \B 0.193 & 0.883 & \B 0.967 & 0.986 & \B 0.996 & \B 0.058 & 0.054 \\
        7 & 128 & 0.732 & \B 0.244 & 0.902 & \B 0.956 & 0.988 & \B 0.995 & \B 0.057 & 0.054 \\
        7 & 256 & 0.956 & \B 0.332 & 0.880 & \B 0.947 & 0.986 & \B 0.993 & \B 0.058 & 0.054 \\
        7 & 512 & 0.931 & \B 0.455 & 0.886 & \B 0.930 & 0.987 & \B 0.991 & \B 0.058 & 0.055 \\
        7 & 798 & 0.982 & \B 0.500 & 0.875 & \B 0.923 & 0.985 & \B 0.991 & \B 0.058 & 0.055 \\
        \bottomrule
      \end{tabular}
    \end{table}
    
    \begin{table}[ht]
      \caption{Results for class 8 inlier between baseline reconstruction error OOD detection approach and our hybrid approach. Best values are bolded.}
      \centering
      \begin{tabular}{llllllllll}
        \toprule
        \multicolumn{2}{c}{ } & \multicolumn{2}{l}{FPR (95\% TPR)} & \multicolumn{2}{l}{AUROC} & \multicolumn{2}{l}{AUPR (in)} & \multicolumn{2}{l}{AUPR (out)} \\
        \cmidrule{3-4}
        \cmidrule{5-6}
        \cmidrule{7-8}
        \cmidrule{9-10}
        Inlier & Bottleneck & Recon. & Hybrid & Recon. & Hybrid & Recon. & Hybrid & Recon. & Hybrid \\
        class & size & err. & (ours) & err. & (ours) & err. & (ours) & err. & (ours) \\
        8 & 2   & 0.999 & \B 0.494 & 0.774 & \B 0.810 & \B 0.972 & 0.970 & \B 0.060 & 0.056 \\
        8 & 4   & 1.000 & \B 0.579 & 0.766 & \B 0.810 & 0.971 & \B 0.973 & \B 0.060 & 0.056 \\
        8 & 8   & 0.998 & \B 0.625 & 0.778 & \B 0.836 & 0.973 & \B 0.978 & \B 0.060 & 0.055 \\
        8 & 16  & 1.000 & \B 0.553 & 0.753 & \B 0.875 & 0.969 & \B 0.984 & \B 0.061 & 0.053 \\
        8 & 32  & 1.000 & \B 0.430 & 0.770 & \B 0.912 & 0.971 & \B 0.989 & \B 0.060 & 0.052 \\
        8 & 64  & 1.000 & \B 0.482 & 0.596 & \B 0.896 & 0.942 & \B 0.986 & \B 0.074 & 0.053 \\
        8 & 128 & 1.000 & \B 0.779 & 0.597 & \B 0.839 & 0.941 & \B 0.978 & \B 0.074 & 0.055 \\
        8 & 256 & 1.000 & \B 0.935 & 0.616 & \B 0.730 & 0.945 & \B 0.963 & \B 0.072 & 0.062 \\
        8 & 512 & 1.000 & \B 0.972 & 0.574 & \B 0.659 & 0.937 & \B 0.953 & \B 0.076 & 0.068 \\
        8 & 798 & 1.000 & \B 0.961 & 0.614 & \B 0.652 & 0.945 & \B 0.949 & \B 0.072 & 0.068 \\
        \bottomrule
      \end{tabular}
    \end{table}
    
    \begin{table}[ht]
      \caption{Results for class 9 inlier between baseline reconstruction error OOD detection approach and our hybrid approach. Best values are bolded.}
      \centering
      \begin{tabular}{llllllllll}
        \toprule
        \multicolumn{2}{c}{ } & \multicolumn{2}{l}{FPR (95\% TPR)} & \multicolumn{2}{l}{AUROC} & \multicolumn{2}{l}{AUPR (in)} & \multicolumn{2}{l}{AUPR (out)} \\
        \cmidrule{3-4}
        \cmidrule{5-6}
        \cmidrule{7-8}
        \cmidrule{9-10}
        Inlier & Bottleneck & Recon. & Hybrid & Recon. & Hybrid & Recon. & Hybrid & Recon. & Hybrid \\
        class & size & err. & (ours) & err. & (ours) & err. & (ours) & err. & (ours) \\
        9 & 2   & 0.592 & \B 0.409 & 0.901 & \B 0.896 & \B 0.988 & 0.985 & \B 0.055 & 0.054 \\
        9 & 4   & 0.603 & \B 0.317 & 0.896 & \B 0.934 & 0.987 & \B 0.991 & \B 0.055 & 0.053 \\
        9 & 8   & 0.506 & \B 0.263 & 0.916 & \B 0.946 & 0.990 & \B 0.993 & \B 0.054 & 0.053 \\
        9 & 16  & 0.660 & \B 0.135 & 0.889 & \B 0.969 & 0.986 & \B 0.996 & \B 0.056 & 0.053 \\
        9 & 32  & 0.814 & \B 0.093 & 0.874 & \B 0.974 & 0.985 & \B 0.996 & \B 0.056 & 0.052 \\
        9 & 64  & 0.721 & \B 0.114 & 0.872 & \B 0.973 & 0.984 & \B 0.996 & \B 0.056 & 0.052 \\
        9 & 128 & 0.625 & \B 0.147 & 0.890 & \B 0.963 & 0.986 & \B 0.995 & \B 0.055 & 0.053 \\
        9 & 256 & 0.873 & \B 0.426 & 0.838 & \B 0.927 & 0.980 & \B 0.990 & \B 0.058 & 0.054 \\
        9 & 512 & 0.912 & \B 0.564 & 0.826 & \B 0.906 & 0.979 & \B 0.988 & \B 0.059 & 0.055 \\
        9 & 798 & 0.979 & \B 0.776 & 0.808 & \B 0.864 & 0.976 & \B 0.983 & \B 0.060 & 0.057 \\
        \bottomrule
      \end{tabular}
    \end{table}

\end{document}